\pdfoutput=1

\documentclass[11pt]{article}

\usepackage[]{acl}

\usepackage{times}
\usepackage{latexsym}

\usepackage{float}
\usepackage{arydshln}
\usepackage{calligra,amsmath,amssymb}
 
\newcommand{\LL}[1]{\textcolor{black}{#1}}
\newcommand{\XZ}[1]{\textcolor{black}{#1}}
\newcommand{\eat}[1]{} 

\usepackage{multirow}
\usepackage{mathrsfs}
\usepackage{graphicx}

\usepackage{comment}
\usepackage[T1]{fontenc}

\usepackage[utf8]{inputenc}

\usepackage{microtype}

\title{Cross-Modal Consistency in Multimodal Large Language Models}

\author{
Xiang Zhang$^{1}$\thanks{$^{*}$Equal contribution.} \qquad Senyu Li$^{2*}$ \qquad Ning Shi$^{2}$ \qquad Bradley Hauer$^{2}$ \qquad Zijun Wu$^{2}$ \\ 
\textbf{Grzegorz Kondrak$^{2}$ \qquad Muhammad Abdul-Mageed$^{1}$ \qquad Laks V.S. Lakshmanan$^{1}$} \\
$^{1}$ University of British Columbia \\
$^{2}$Alberta Machine Intelligence Institute, Dept. of Computing Science, University of Alberta \\
\texttt{\{wyattz23,muhammad.mageed,laks\}@cs.ubc.ca}\\
\texttt{
\{xzhang23,senyu,ning.shi,bhauer,zijun4,gkondrak\}@ualberta.ca} 
}

\begin{document}

\maketitle

\begin{abstract}
Recent developments in multimodal methodologies have marked the beginning of an exciting era for models adept at processing diverse data types, encompassing text, audio, and visual content. Models like GPT-4V, which merge computer vision with advanced language processing, exhibit extraordinary proficiency in handling intricate tasks that require a simultaneous understanding of both textual and visual information. Prior research efforts have meticulously evaluated the efficacy of these Vision Large Language Models (VLLMs) in various domains, including object detection, image captioning, and other related fields. However, existing analyses have often suffered from limitations, primarily centering on the isolated evaluation of each modality's performance while neglecting to explore their intricate cross-modal interactions. Specifically, the question of whether these models achieve the same level of accuracy when confronted with identical task instances across different modalities remains unanswered. In this study, we take the initiative to delve into the interaction and comparison among these modalities of interest by introducing a novel concept termed {\em{cross-modal consistency}}. Furthermore, we propose a quantitative evaluation framework founded on this concept.   Our experimental findings, drawn from a curated collection of parallel vision-language datasets developed by us, unveil a pronounced inconsistency between the vision and language modalities within GPT-4V, despite its portrayal as a unified multimodal model. Our research yields insights into the appropriate utilization of such models and hints at potential avenues for enhancing their design.
\end{abstract}

\section{Introduction}
\begin{figure}[t!]
    \centering
    \includegraphics[width=0.9\columnwidth]{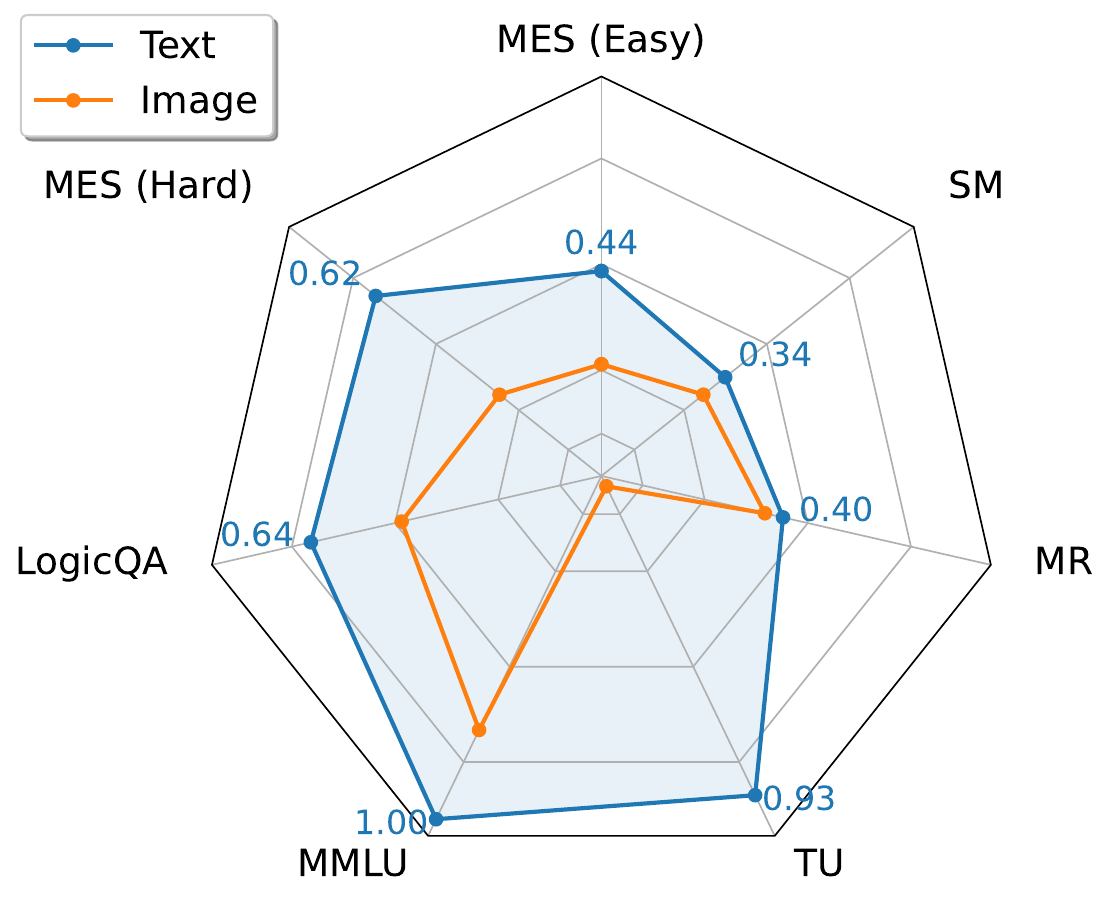}
    \caption{
  Visualization of the performance gap between the modality of text and image in seven different tasks. 
    }
    \label{fig:spider}
\end{figure}

Recent large multimodal models have showcased remarkable capabilities in tasks that require the integration of multiple modalities and sources of information~\cite{huang2023chatgpt}. Among these, the performance of Vision Large Language Models (VLLMs)~\cite{zhang2023vision,yang2023dawn} stands out, thanks to the vast amounts of image and text data available for training and the rapid progress in both computer vision and language modelling.
However, due to the distinct training methodologies employed by these models, such as contrastive learning~\cite{radford2021learning,jin2024contranovo} and embodied image-language modeling~\cite{driess2023palme}, and the varying quality of training data for each modality~\cite{yin2023ttidacontrollablegenerativedata}, these networks often exhibit performance disparities across different modalities.

Previous research has extensively evaluated the performance of individual modalities in multimodal systems. For instance, ~\newcite{yang2023dawn} conducted a thorough assessment of GPT-4V's vision understanding capabilities, and~\newcite{chen2023endtoend} analyzed model's decision-making abilities. However, assessing a model's performance on each individual modality in isolation does not fully evaluate its true multimodal abilities. It is possible, for example, for a model to excel in numerous vision tasks but still lag significantly behind in language understanding.  Moreover, simply testing performance on individual tasks provides no insight into whether and how each modality of the model influences the others. Unfortunately, the cross-modality relationship is frequently overlooked in the aforementioned research.

In this study, we go beyond the traditional approach of simply evaluating multimodal systems through separate downstream tasks and reporting their scores. Our focus is primarily on measuring the inherent \textit{differences} in capabilities between various modalities, with special attention to vision and language, given their prominence among other modalities. To enable a comprehensive analysis, we introduce the concept of {\em cross-modal consistency}, complete with a formal definition and an evaluation framework. We consider cross-modal  consistency to be an essential element in the design of complex multimodal systems with neural components, as it guarantees coherence and reliability in the system's performance. This is crucial for both interpretability and for fostering user trust.

We subsequently construct a comprehensive vision-language parallel dataset encompassing seven tasks, each designed to highlight different facets of vision and language capabilities. This dataset serves as a tool for evaluating the vision-language consistency of  VLLMs. Our experiments with the GPT-4V model on the dataset reveal significant  inconsistencies between its vision and language capabilities. The results indicate that its performance varies considerably depending on whether the same task instance is prompted in one modality versus the other.


Our contributions are: 
(1)~We introduce the novel concept of cross-modal consistency, along with a comprehensive evaluation framework. This approach transcends traditional assessment methods for multimodal models, which typically evaluate each modality in isolation. 
(2)~We develop and release seven diverse datasets, carefully designed for vision-language consistency evaluation, opening up opportunities to exploit these datasets in future research. 
(3)~Our experiments on GPT-4V reveal a significant disparity between vision and language abilities within such a system, prompting the introduction of the Vision-Depicting-Prompting (VDP) method as a potential remedy. Our findings offer valuable guidance for more effective future use of such multimodal models.

\section{Related Work}


\XZ{A substantial amount of effort has been dedicated to meticulous evaluation of large multimodal models such as GPT-4V. To  assess the capabilities of these models across all their modalities, a wide array of tasks has been  tested. E.g., researchers have scrutinized GPT-4V's aptitude in solving problems within specialized domains, including biomedicine~\cite{liu2023holistic}, medical applications~\cite{wu2023gpt4vision}, and autonomous driving~\cite{wen2023road}, employing intricate image inputs. Beyond these domain-specific evaluations, more general skills like chart image understanding~\cite{liu2023mmc} and optical character recognition~\cite{shi2023exploring} have also been analyzed. However, these evaluations often focus solely on performance metrics for each test dataset, with little or no exploration of the relative capability gaps between vision and language. In this study, our primary emphasis lies in uncovering the relative \textit{disparities} in the abilities of multimodal models across their various modalities, rather than merely assessing absolute performance within specific tasks.}

Despite the lack of cross-modal analysis for multimodal models, previous research has delved into examining cross-lingual abilities in Large Language Models (LLMs). For example, by translating task instances into different languages and analyzing the pairwise results, ~\newcite{zhang2023don} demonstrated that models like GPT-3.5, primarily trained on English text corpora, exhibit disparities in their performance across various tasks when prompted with different languages. Specifically, these LLMs display a bias toward English. Also, ~\newcite{chou2024mm} analyze the consistency concept \textit{within} each modality for multi-modal models. We extend our research to encompass consistency analysis across various modalities, recognizing that different languages can be regarded as distinct modalities as well. Our generalized framework  sheds light on the underlying principles governing the {\bf consistency}  of multimodal models when confronted with \LL{tasks in} diverse modalities, thereby contributing to a deeper understanding of their capabilities and limitations.

\section{Preliminaries and Key Concepts}
As ``consistency'' can carry different interpretations within the specific context we are addressing, a formal definition of the concept of cross-modal consistency for multimodal models is warranted. To that end, we establish an instance of task $t$, represented as the paired value $(d_a, q)$. Here, $d_a$ represents a \LL{data}  element from the input space $\mathcal{D}_a$ corresponding to modality $a$, while $q \in \mathcal{Q}$ represents the abstract query, often presented in the form of a question pertinent to the task at hand. A task set within modality $a$ is then constituted by combining certain \LL{data} elements from modality $a$ with the queries $q$, which can be denoted as $S_{t,a} = \{(d^{(1)}_a, q), (d^{(2)}_a, q), (d^{(3)}_a, q), \ldots \}$. When the queries $q$ are held constant, and elements $d_b \in \mathcal{D}_b$ \LL{in another modality $b$} are gathered, we obtain the corresponding task set in another modality, denoted as $S_{q,b}$. In essence, the task $t$ embodies the task-specific queries, encompassing, e.g., activities such as solving equations, translation, question answering, etc. Meanwhile, the data elements $d_m$ may take the form of equation instances or question descriptions within modality $m$, which can \LL{involve the modalities of} image, text, or speech. 

\XZ{We  introduce the concept of a 'converter,'  a function $K_{a,b}: \mathcal{D}_a \mapsto \mathcal{D}_b$ which maps data elements from modality $a$ to  $b$. While there exist various methods for converting data between modalities (e.g., from language to vision through taking a picture), we are specifically interested in converters that preserve information necessary for solving a given task with query $q$, denoted as $K^{q}_{a,b}$. Information-preserving converters are distinctive, as the correct answer for a given task instance $(d, q)$ depends solely on the information within $d$ rather than its modality. Therefore, both $(d_a, q)$ and $(K^q_{a,b}(d_a), q)$ are guaranteed to share the same gold label.
In this work, we assume the existence of $K^q$ for every $q \in \mathcal{Q}$, but  finding such a converter is beyond the scope of this paper.  Inter-modality conversion may be challenging for certain modalities. Some tasks may involve aspects of information, such as emotions in speech or nuanced visual perception in  images, that cannot be easily preserved during conversion. We design our experiments  with tasks where a  $K^q$ clearly exists.
}

A multimodal model can be conceptualized as a function, denoted $M: \mathcal{D} \times \mathcal{Q} \mapsto \mathcal{Y}$, \LL{mapping data elements and queries to an answer}.  Here, $\mathcal{D}$ represents the collective space encompassing all the modalities of interest, formally  $\mathcal{D} = \bigcup\limits_{m} D_m$, where $m$ spans over all relevant modalities. On the other hand, \LL{the answer space} $\mathcal{Y}$ refers to a unified and structured representation, which, in the case of GPT-4V, assumes the form of text.

A model $M$ is said to exhibit consistency between modalities $a$ and $b$ provided:
\begin{equation*}
    M(d_a, q) = M(K^{q}_{a,b}(d_a), q), 
    \forall d_a \in D_a, 
    \ q \in \mathcal{Q}
\end{equation*}
In other words, $M$ is \textit{consistent} if its output is invariant 
under any modality transformation $K^q$ which preserves
all essential information necessary for solving the task associated with query $q$. 
E.g.,
consider  solving mathematical equations.
A model which solves this task is consistent across the text and image modalities
if neither transcribing the equation from  image to text,
nor imaging an equation presented as text,
changes the model's output.

In short, a consistent model should remain agnostic to the modality of the task instance and yield identical results as long as an equivalent amount of information is provided, reflecting its capacity to handle multimodal data seamlessly.


\section{Method}

\begin{figure}[t!]
    \centering
    \includegraphics[width=\columnwidth]{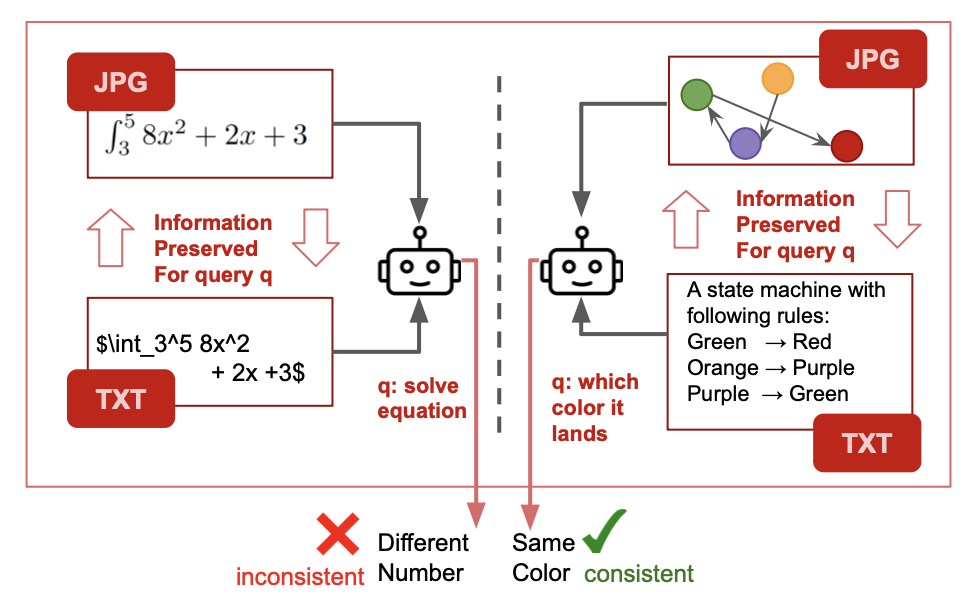}
    \caption{
  Illustration of the concept of cross-modal consistency. A consistent model (right) applies the same internal reasoning to task instances with identical information, regardless of the encoding modality, leading to consistent outcomes. In contrast, an inconsistent model displays significant behavioral changes in response to different input modalities, resulting in varying outcomes as the modality alters.
    }
    \label{fig:comp}
\end{figure}

In this section, we describe our method for testing cross-modal consistency.
We establish a quantitative evaluation framework, 
with a focus on the vision-language cross-modality. 
We provide a description of our methodology
and the specific metrics we propose for evaluation.

\subsection{WorkFlow}
\label{sec:method}
For an instance set of a given task $t$ in modality $a$, denoted as $S_{t,a}$ = $\{(d^{(1)}_a, q)$, $(d^{(2)}_a, q)$, $(d^{(3)}_a, q)$, $\cdots\}$, our first step involves constructing a parallel instance set $S_{t,b}$ in modality $b$ using an information-preserving converter $K^q_{a,b}$. \XZ{We do so by applying  $K^q_{a,b}$ to each data object $d^{(i)}_a$ to get the object $d^{(i)}_b := K^q_{a,b}(d_a)$  in modality $b$. By doing so, each paired instance $(d^{(i)}_a, q)$ and $(d^{(i)}_b, q)$ shares the same gold label since the information in $d$ is preserved for the task with query $q$.
 }
In the context of analyzing the vision and language modalities, our converter 
is comprised of an optical character recognition (OCR) system combined with human verification for converting images to text, and screenshot software for converting text into images. We carefully select tasks where the information required for solving the task can be fully retained through the utilization of this converter, as exemplified by mathematical equation solving.

Next, we independently apply the  model $M$ to each pair of instances $(d^{(i)}_a, q)$ and $(d^{(i)}_b, q)$ to obtain pairwise results $M(d^{(i)}_a, q)$ and $M(d^{(i)}_b, q)$. 

\subsection{Metrics}
We  introduce our task consistency score $C_t$ based on these pairwise instances:

\begin{equation}
C_t = \frac{1}{n}\sum_{i=1}^{n} c^i_M
\end{equation}

where

\begin{equation}
c^i_M=\begin{cases}
1, & \text{if } M(d^{(i)}_a, q) = M(d^{(i)}_b, q)\\
0, & \text{otherwise}
\end{cases}
\end{equation}

In essence, $C_t$ is the proportion of instances for which  \LL{model $M$ has consistent  performance on the given task, between modalities $a$ and $b$}.

\section{Experiments}\label{sec:exps}
\begin{figure*}[t!]
    \centering
    \includegraphics[width=0.9\linewidth]{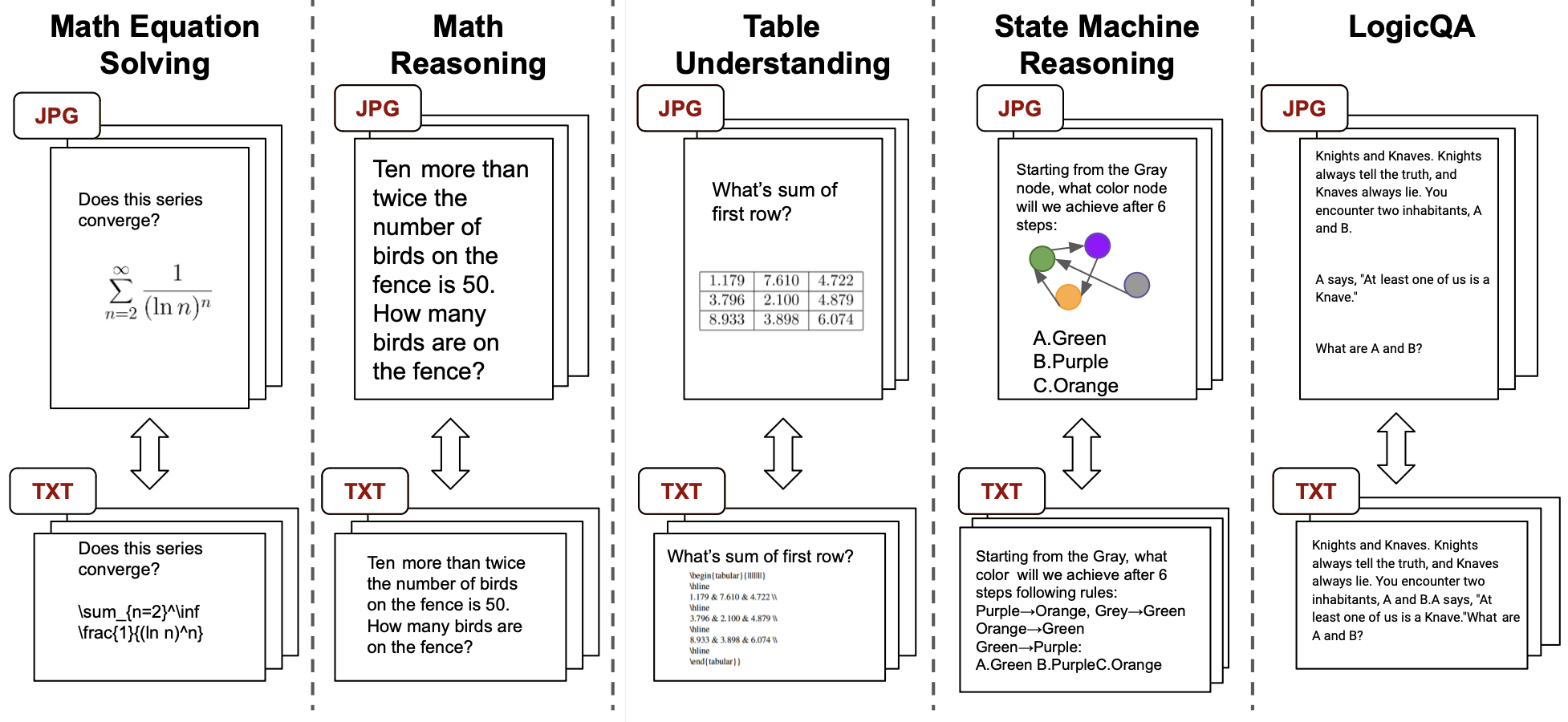}
    \caption{An Overview of the Components of Our Vision-Language Consistency Dataset. Data instances are presented in pairs, featuring one in the vision modality and another in the text modality. Notably,  Math Equation Solving dataset encompasses two segments, each representing different difficulty levels. }
    \label{fig:Dataset}
\end{figure*}

\subsection{Data Construction}
\label{subsec:dataset}
Since there is currently no existing parallel vision-language task dataset, we \LL{create our own datasets for both our experiments and also to facilitate future research endeavors.} Following the approach outlined in Section \ref{sec:method}, we meticulously selected seven tasks that gauge various facets of Vision-Large Language models. For each of these tasks, we ensure that data instances can be transformed between image and text formats while preserving all task-related information, utilizing a straightforward converter (e.g., OCR). Recognizing that a flawless converter does not exist in practice, we undertake the manual verification of each converted data instance to prevent any potential errors during the conversion process. We will make our dataset available for use by the research community in the final version of our paper.

\subsubsection{Task Description.}
\textbf{Math Equation Solving.} Mathematical reasoning stands as a cornerstone of multi-modal models' capabilities. Mathematical problems typically involve equations presented in a visual format, offering a clear depiction of intricate symbols and notations. Given that formulas can be seamlessly converted to text formats like LaTeX without losing any essential information for solving these equations, constructing a parallel dataset for such tasks is a natural fit for analyzing cross-modal consistency. For our dataset, we source math questions with equations from two distinct origins, each representing varying levels of difficulty. For low difficulty levels, we extract 901 high school-level mathematical questions in LaTeX (text) format from MATH dataset ~\cite{hendrycks2021measuring1}, rendering each question using a LaTeX compiler to generate corresponding image data. To introduce a greater level of complexity, we gathered 50 college-level calculus questions, along with their corresponding answers, using the same procedure. Consequently, we paired all the image-based math questions with their corresponding text representations to create our comprehensive equation-solving dataset, encompassing both easy and challenging questions. An illustrative example of this dataset can be found in Figure \ref{fig:Dataset}, and detailed data samples are available in Appendix \ref{app:MESE} and  Appendix \ref{app:MESH} . \\
\textbf{Logical Reasoning.} To assess the vision-language consistency in logical reasoning abilities for the VLLMs, we employ two distinct datasets: GSM8K~\cite{DBLP:journals/corr/abs-2110-14168} and LogicQA~\cite{liu2020logiqa}.
GSM8K comprises 8,500 question instances in text format, with each instance representing a problem description in English text paired with a labeled answer. We transform the text into images by capturing screenshots of the rendered text with an appropriate font size and layout.
Similarly, LogicQA consists of 8,678 more challenging questions presented in text format and each is converted into an image by us. Subsequently, we pair these resulting images with the original text files, creating a parallel dataset that enables the exploration of this task in both image and text modalities. An illustrative example of this dataset construction can be found in Figure \ref{fig:Dataset}, and detailed data samples are available in Appendix \ref{app:MR} and Appendix \ref{app:LogiQA}.

\textbf{Table Understanding.} Tables, commonly encountered in everyday life, are often presented as images, and the effective extraction of information from them is vital for various tasks. As well-structured table images can be easily converted into LaTeX text, they serve as an excellent choice for conducting vision-language consistency analysis.
To facilitate this analysis, we creat 30 distinct tables in LaTeX, each featuring multiple rows and columns, with numerical values in each cell. Our task revolves around accurately summing the numbers within a given row and column. We provide parallel task instances in both LaTeX text and rendered images, as illustrated in Appendix \ref{app:TU} 

\textbf{State Machine Reasoning.} State machines, which can be effectively visualized as graphs or represented through text with transition rules, serve as an ideal test bed for vision-language consistency in \LL{simple} computational capabilities of VLLMs. Our approach involves generating images of state machines with varying total numbers of nodes (states). Each node in the state machine is assigned a distinct color and features precisely one outgoing edge, ensuring a unique path and solution.
The questions we formulate are of the form, "Starting from the color grey, after \textit{n} steps, which color will we end up in?" Here, n is a variable that we select. Additionally, we generate a text version of these state machines by listing out all the transition rules corresponding to the arrows. To prevent any form of cheating by looking at the last state in the text, we shuffle the order of the rules. We create state machines with different numbers of states and questions with varying numbers of steps, to introduce varying difficulty levels. The data samples can be seen in Appendix \ref{app:state}. 

\textbf{Reading Comprehension.} To assess the model's consistency in comprehending lengthy English paragraphs across vision and language modalities, we provide the model with the same text content in two different formats: plain text and images of the text. We employ the test part of the Massive Multitask Language Understanding ~\cite{hendrycks2021measuring2}, or MMLU dataset as our source, which includes 1,477 extensive text passages, each accompanied by multiple-choice questions designed to evaluate the comprehension of the text content.
For this dataset, we convert each text instance into an image by rendering the text into a PDF before converting it to a JPG image. Detailed data samples can be found in the Appendix \ref{app:MMLU}.\\

\subsection{Experiment Details.}
\begin{table}
    \centering
    \resizebox{0.44\textwidth}{!}{
    \begin{tabular}{|cccc|}
    \hline
    \textbf{Task} & \textbf{Modal} & \textbf{Acc} & \textbf{Consistency} \\
    \hline\hline
     \multirow{2}{*}{MES(Easy)} & Text & 0.44 & \multirow{2}{*}{0.72} \\
    \cdashline{2-3}
     & Image & \ \ \ \ 0.24 $\color{red}\Downarrow$ &\\
    \hline\cline{1-4}
    \multirow{2}{*}{MES (Hard)} & Text & 0.62 & \multirow{2}{*}{0.62} \\
    \cdashline{2-3}
      & Image & \ \ \ \ 0.28 $\color{red}\Downarrow$ & \\
    
    \hline\cline{1-4}
    \multirow{2}{*}{LogicQA} & Text & 0.64 & \multirow{2}{*}{0.64} \\
    \cdashline{2-3}
      & Image & \ \ \ \ 0.44 $\color{red}\Downarrow$ &  \\
    \hline\cline{1-4}
     \multirow{2}{*}{MMLU} & Text & 1.00 & \multirow{2}{*}{0.74} \\
    \cdashline{2-3}
      & Image & \ \ \ \ 0.74 $\color{red}\Downarrow$ &  \\
    \hline\cline{1-4}
    
    \multirow{2}{*}{TU} & Text & 0.93 & \multirow{2}{*}{0.10} \\
    \cdashline{2-3}
     & Image & \ \ \ \  0.03 $\color{red}\Downarrow$ & \\
   
    \hline\cline{1-4}
    \multirow{2}{*}{ MR} & Text & 0.40 & \multirow{2}{*}{0.92} \\
    \cdashline{2-3}
    & Image &  0.36 &  \\
    \hline\cline{1-4} 
    \multirow{2}{*}{State Machine} & Text & 0.34 & \multirow{2}{*}{0.67} \\
    \cdashline{2-3}
      & Image & 0.28 &  \\

    \hline
\end{tabular}}
    \caption{Test results for vision-language consistency datasets.
    MES stands for Math Equation Solving, TU stands for Table Understanding and MR stands for math reasoning. The symbol $\color{red}\Downarrow$ denotes a sizeable decrease in accuracy (greater than 10\%) when input is in the image format. }
    \label{tab:main}
\end{table}
We apply our framework and constructed \LL{datasets} to evaluate the cross-modal consistency of the OpenAI GPT-4V model, known for its proficiency in both vision and language modalities. Given the limited daily access to prompt this model, our experiments were conducted on a randomly selected subset of 50 samples from each dataset.  We select the GPT-4V classical mode, which does not include additional plug-ins and employs a relatively low decoding temperature to minimize \LL{variance} in its output. To ensure a fair comparison of capabilities between the two modalities, we embedded the query questions into the image and exclusively used images for prompting. This avoids the involvement of any text input  when testing the vision modality. Additionally, to prevent the model from performing reasoning steps in text and introducing unintended modality conversions, we explicitly instructed the model to output answers without any reasoning steps. Our results are manually collected for pairwise data instances, and we calculate the consistency scores based on \LL{the methodology outlined in Section~\ref{sec:method}.} 
\subsection{Main Results}
\label{sec:res}

The main outcomes of our assessments across seven distinct datasets are outlined in Table \ref{tab:main}. Notably, even though the input contains an equivalent amount of information necessary for task completion, substantial disparities emerge between image and text input formats. This phenomenon occurs even in tasks where images are conventionally considered to offer a more vivid and intuitive representation from a human perspective. 

\LL{We note that consistency, being based on response agreement between modalities, can be high or low regardless of per modality accuracy. The highest consistency (0.92) is observed for math reasoning even though both modalities have a relatively low accuracy ($\leq 0.40$). By contrast, the consistency drops to 0.64 on logical reasoning (LogicQA) on which the individual modalities have higher accuracy ($\geq 0.44$).}

For tasks that involve intricate reasoning steps, including equation solving, math/logical reasoning, and state machine reasoning, we observe relatively low accuracy even when the input is presented in pure text format. These tasks align with areas where the model generally struggles. When the input modality shifts to using images, the proficiency in solving such tasks deteriorates further, resulting in a noticeable drop in performance, despite the fact that the images contain an equal amount of information. This emphasizes the substantial inconsistency in task-solving across modalities and highlights the model's superior ability in one modality (Language) compared to the other (Vision).

On the other hand, for tasks primarily focused on extracting information from provided content and comprehending that information, such as Language Understanding and Table Understanding, we witness near-perfect performance when the model is prompted with text input. However, a more significant drop in accuracy (up to 90\%) is observed in such tasks when the input modality shifts to images. This indicates that the change in modality significantly \LL{impacts} the model's processing capabilities, providing strong evidence of the inconsistency of \LL{the model}.



\begin{figure*}[t!]
    \centering
    \includegraphics[width=0.7\linewidth]{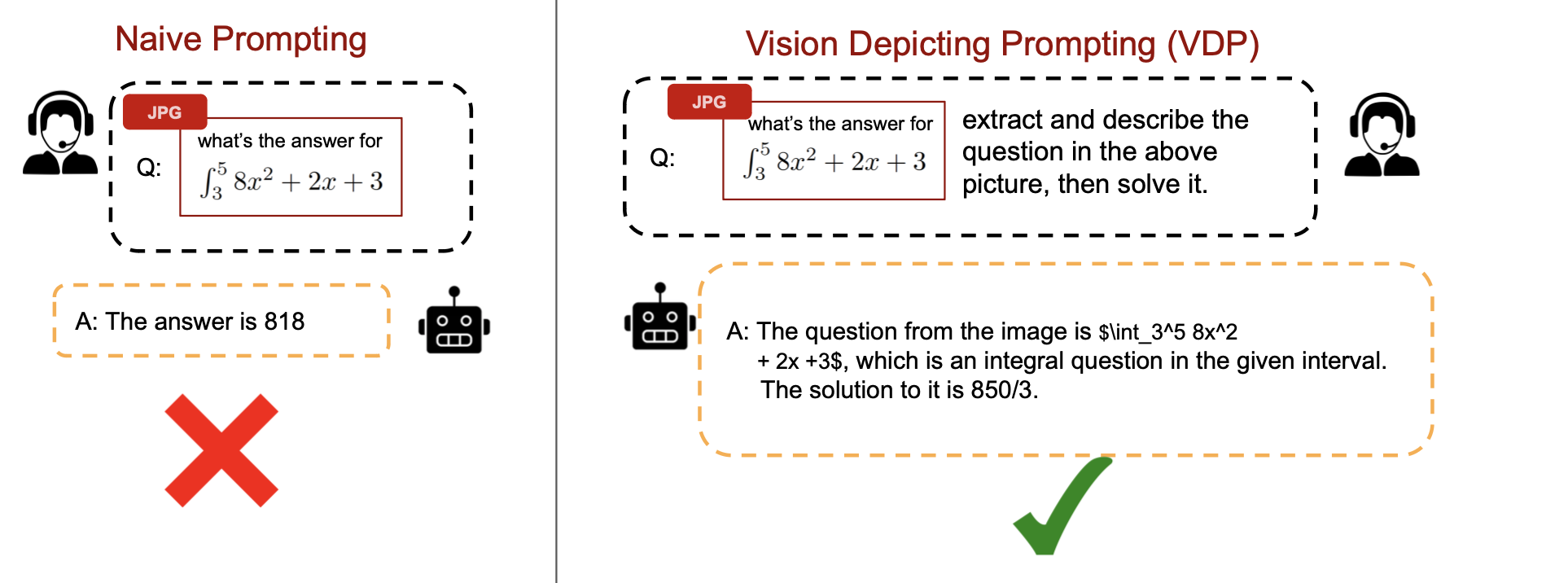}
    \caption{Overview of the VDP Method: The left part illustrates the conventional approach to prompting vision tasks, while the right part demonstrates VDP in comparison. }
    \label{fig:VDP}
\end{figure*}

In conclusion, in multimodal systems like GPT-4V, the language modality demonstrates a \textbf{dominant} advantage \LL{over vision modality,} when tasks are tackled in text format, despite the presence of the same information in image format. This strongly suggests a {\it non-consistent} cross-modal behavior within the network. \LL{While each modality exhibits varying levels of task-solving and reasoning capabilities, the inconsistency across modalities is observed across tasks regardless of the accuracy level of each modality for the task in hand.  }


\subsection{Ablation Study on Content Extraction from Images}
As solving tasks in image format inevitably requires accessing essential information from the images, we conducted additional experiments to investigate whether the performance gap is attributable to the model's inability to access information. To address this, we conducted one-step Optical Character Recognition (OCR) using the model's own network on all instances of tasks that exhibited a significant performance gap  between image and text. Specifically, for each image input (indicated by the red arrow in Table \ref{tab:main}), we prompt the model with the instruction 'extract the exact content in the image' and compare the results with the original input to determine if they match. This approach allows us to eliminate the possibility that the performance issues in image format are due to the model's inability to correctly recognize the input.

As shown in Table \ref{tab:OCR}, OCR accuracy approaches nearly 100\% for all instances of LogicQA, MMLU, and Table Understanding tasks. This suggests that the model faces no difficulties in accurately extracting information, such as numbers from each row and column in table images. The substantial gap (up to 90\%) in accuracy (Table \ref{tab:main}) between images and text can be attributed solely to the model's internal reasoning processes for each modality. This underscores the inconsistent internal reasoning employed by the model when presented with the same content in different modalities.  

In contrast, we observe lower OCR accuracy for Math equation-solving inputs, as complex math equations pose challenges for accurate recognition and extraction. To \LL{isolate and} distinguish the source of inconsistency  --  \LL{inaccurate recognition of image data \textit{or} \eat{those stemming from} poor  actual internal reasoning,}  we report \textit{conditional consistency scores} for image instances given correct versus incorrect OCR results. From Table \ref{tab:OCRConsistency}, it becomes evident that there is no direct correlation between consistency scores and direct OCR accuracy. This further bolsters our claim that such models simply exhibit distinct \LL{(and inconsistent!)} internal behaviors under different modalities.
\begin{table}
\centering
\resizebox{0.68\linewidth}{!}{%
\begin{tabular}{c|c}
\hline
\textbf{DataSet} &  \textbf{OCR Accuracy}  \\
\hline\hline
 MES (Easy) & 0.68  \\ 

MES (Hard)   & 0.76  \\ 

LogicQA  & 0.98   \\ 

MMLU & 0.98  \\ 

TU  & 1.00 \\ 
\hline
\end{tabular}
}

\caption{Result of performing OCR on the all images of experimented task instances.}
\label{tab:OCR}
\end{table}

\begin{table}
\centering
\resizebox{0.78\linewidth}{!}{%
\begin{tabular}{c|cc}
\hline
\textbf{DataSet} &  \textbf{YConsistency}  &  \textbf{NConsistency} \\
\hline\hline
 MES (Easy) & 0.70 & 0.75 \\ 

MES (Hard)   & 0.66 & 0.58 \\ 

\hline
\end{tabular}
}

\caption{Conditional vision-language consistency  score given the OCR results. The term 'YConsistency' refers to the consistency given OCR outputs are correct. Conversely, 'NConsistency' denotes the consistency score given incorrect OCR outputs. }
\label{tab:OCRConsistency}
\end{table}

\label{sec:abl}



\section{Vision-Depicting-Prompting (VDP) }
As shown in Section~\ref{sec:res}, for the same \LL{task}, VLLMs such as GPT-4V perform much better when questions are presented in text format, even when the information can be completely extracted  from the image instances. \LL{Inspired by these findings, we propose a novel method of \textit{Vision-depicting-prompting} (VDP) for improving model's reasoning ability through image context. We now explain VDP. } 
\subsection{Prompting Details}
In the case of a task instance presented in image format, VDP diverges from directly soliciting an answer solely based on the image input, as illustrated in Figure~\ref{fig:VDP}. Instead, we adopt a two-step process: we first prompt the model to extract and articulate the description of the image task using textual language. This aims to maximize the transformation of the image signal into a text signal, recognizing the inherently stronger reasoning abilities associated with text information, as demonstrated earlier. Subsequently, we prompt the model to provide an answer, taking into account both the text description of the task and the original image input, as depicted in Figure~\ref{fig:VDP}. 

Unlike previous research that sought to enhance the reasoning abilities of multimodal models by augmenting input images with supplementary text~\cite{lin2022revive,hu2023promptcap,zhang-etal-2023-bridging}, VDP does not focus on information augmentation. Particularly in the task instances designed for our study, images already contain all the necessary information required to complete the task. Therefore, converting these images into text format does not provide any additional information that aids in solving the task. Instead, VDP is rooted in the observation that textual signals can significantly stimulate a model's reasoning capability as model has a bias towards language modality. Instead, VDP is based on the observation that textual signals can significantly stimulate a model's reasoning capability, given the model's inherent bias toward the language modality. VDP achieves this by explicitly extracting textual information from the images, thus directly leveraging the model's language processing capabilities more effectively.  

\begin{table}
\centering
\resizebox{\linewidth}{!}{%
\begin{tabular}{c|cccc}
\hline
\textbf{Task} & \textbf{Modality} & \textbf{Prompt} & \textbf{Acc} & \textbf{Consistency} \\
\hline\hline
 \multirow{3}{*}{MES (Easy)}& text & naive & 0.44 & ---- \\ 

 & \multirow{2}{*}{image} & naive  & 0.24 & 0.72\\

 & & VDP & \textbf{ \ \ 0.48 $\Uparrow$} & 0.72  \\
\hline

\multirow{3}{*}{MES (Hard)}& text & naive &   0.62  & ---- \\ 

 & \multirow{2}{*}{image} & naive  & 0.28 & 0.62\\

 & & VDP &  \textbf{\ \ \ \ 0.50 $\Uparrow$ } & \textbf{ \ \ 0.76 $\Uparrow$} \\
\hline

\hline

\multirow{3}{*}{LogicQA}& text & naive &   0.64  & ---- \\ 

 & \multirow{2}{*}{image} & naive  & 0.44 & 0.64\\

 & & VDP &  \textbf{\ \ \ \ 0.56 $\Uparrow$ } & \textbf{ \ \ 0.80 $\Uparrow$} \\
\hline

\hline

\multirow{3}{*}{MMLU}& text & naive &   1.00  & ---- \\ 

 & \multirow{2}{*}{image} & naive  & 0.74 & 0.74\\

 & & VDP &  \textbf{\ \ \ \ 0.98 $\Uparrow$ } & \textbf{ \ \ 0.98 $\Uparrow$} \\
\hline

\hline

\multirow{3}{*}{TU}& text & naive &   0.93  & ---- \\ 

 & \multirow{2}{*}{image} & naive  & 0.03 & 0.10\\

 & & VDP &  \textbf{\ \ \ \ 0.93 $\Uparrow$ } & \textbf{ \ \ 0.90 $\Uparrow$} \\
\hline

\end{tabular}
}

\caption{Result of VDP prompting. MES stands for Math Equation Solving and TU stands for Table  Understanding. $\Uparrow$ represents an improvement of more than 10\% using VDP. }
\label{tab:VDP}
\end{table}

\subsection{Experiment Results for VDP}
We apply VDP to five of the tasks previously examined in Section~\ref{sec:exps}, where these tasks demonstrate notable performance disparities between image and text inputs. We therefore investigate whether VDP can effectively bridge the performance gap between modalities on such tasks. The outcomes are detailed in Table~\ref{tab:VDP}. 

Remarkably, we  observe a substantial improvement in accuracy exceeding 12\% when solving problems within the realm of vision modalities using VDP, as compared to naive prompting. In tasks requiring reasoning abilities, we note an average accuracy enhancement of 19\%. However, the overall performance still lags behind that of text-based prompting. This discrepancy can likely be attributed to the challenges in accurately depicting and extracting information from objects within images during VDP.
In contrast, an impressive average increase of 57\% in accuracy is observed in tasks centered around understanding (TU and MMLU). Particularly, in the case of table understanding, we  witness a remarkable 90\% boost in accuracy, particularly when the table's content is extracted before any necessary calculations are applied. For these tasks, we find that performance eventually reaches parity with text-based prompting, underscoring the effectiveness of VDP, particularly in tasks that involve a deeper understanding of the information within the input instances. 

\LL{Furthermore, there is a substantial increase in the consistency score with VDP compared to prompting with plain images \LL{(naive prompting), e.g., from 0.64 to 0.80 on LogicQA and from 0.10 to 0.90 on TU.} These results reinforce our hypothesis that models such as GPT-4V exhibit varied and often inconsistent reasoning capabilities across different modalities and underscore the effectiveness of our VDP approach for enhancing consistency. Properly addressing such disparities between modalities as done by our VDP approach can also help to improve the performance in solving the tasks.}  

\section{Conclusion}
In this study, we performed a systematic analysis of the consistency across modalities in multimodal systems. Our results demonstrate that models such as GPT-4V maintain a relatively independent internal representation of reasoning between visual and textual signals, as evidenced by results we obtained on our datasets which we  specially designed for the tasks. Notably, GPT-4V exhibits superior performance in language modeling compared to reasoning within a visual context. These findings offer valuable insights into the potential applications of such multimodal systems and highlight the need for more integrated system designs. Furthermore, we introduce a Vision-depicting-Prompting solution to effectively address this inconsistency.

\section*{Limitations}
While our method is straightforward and effective in revealing inconsistency across modalities, it does encounter challenges when applied to certain existing tasks. Obtaining an information-preserving converter from one modality to another can prove difficult for specific tasks, such as detecting emotions from speech. Consequently, we may not always be able to readily convert the modality of every given dataset and evaluate the cross-modal consistency of these tasks. However, it is important to note that this limitation should not undermine the value of our approach. Our method provides a general framework for assessing cross-modal behavior, and there exist numerous tasks that can be easily converted across modalities without any loss of information, as demonstrated in our \LL{constructed} datasets. By testing on such tasks, we can gain a comprehensive understanding of a model's cross-modal behavior.

\section*{Ethical Consideration}
Our exploration of modality consistency serves as a valuable means to enhance the transparency of multimodal models and gain a profound comprehension of their behavior. By delving into the alignment of model responses across diverse modalities, we uncover intricate insights into the decision-making processes and rationale behind their actions. This comprehensive understanding not only instills confidence in the outcomes generated by these models but also significantly enhances their overall interpretability. Transparency in this context becomes essential not only for establishing trust when these models are integral to pivotal decision-making processes but also for addressing ethical and societal implications. As we unravel the intricacies of multimodal reasoning, it underscores the necessity for continuous ethical contemplation and the implementation of proactive measures to address potential challenges arising from advanced multimodal models.

\bibliography{custom}
\bibliographystyle{acl_natbib}

\clearpage
\onecolumn  
\section*{}
\

\appendix
\label{sec:}

\section{Math Equation Solving (Easy) Dataset}
\label{app:MESE}

\begin{figure*}[ht]
    \centering
    \includegraphics[width=0.7\linewidth]{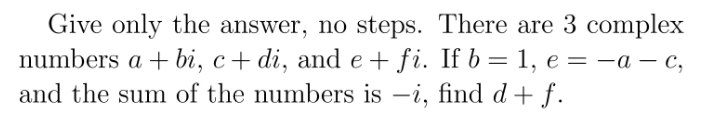}
    \caption{Sample 1 of Math Equation Solving (Easy) Dataset: Image.  }
    \label{fig:MESE1}
\end{figure*}

\begin{table*}[ht]
    \centering
    
    \resizebox{\textwidth}{!}
    {%
    \begin{tabular}{|lp{1\linewidth}|}

    \hline
      \textbf{Text:}&
\texttt Give only the answer, no steps. Find the largest value of \$c\$ such that \$1\$ is in the range of \$f(x)=x \^ \ 2-5x+c\$.
    \\ 
   \hline \end{tabular} 
    }

    \caption{Sample 1 of Math Equation Solving (Easy) Dataset: Text  
    }
    \label{table:MESE1}
    
\end{table*}

\clearpage

\begin{figure*}[ht]
    \centering
    \includegraphics[width=0.72\linewidth]{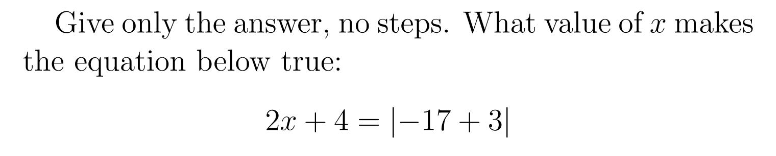}
    \caption{Sample 2 of Math Equation Solving (Easy) Dataset: Image.  }
    \label{fig:MESE2}
\end{figure*}

\begin{table*}[h]
    \centering
    
    \resizebox{0.95\textwidth}{!}
    {%
    \begin{tabular}{|lp{1\linewidth}|}

    \hline
      \textbf{Text:}&
\texttt  Give only the answer, no steps. What value of $x$ makes the equation below true: \$\$2x + 4 = |{-17 + 3}|\$\$
    \\ 
   \hline \end{tabular} 
    }

    \caption{Sample 2 of Math Equation Solving (Easy) Dataset: Text  
    }
    \label{table:MESE2}
    
\end{table*}

\newpage

\section{Math Equation Solving (Hard) Dataset}
\label{app:MESH}

\begin{figure*}[ht]
    \centering
    \includegraphics[width=0.7\linewidth]{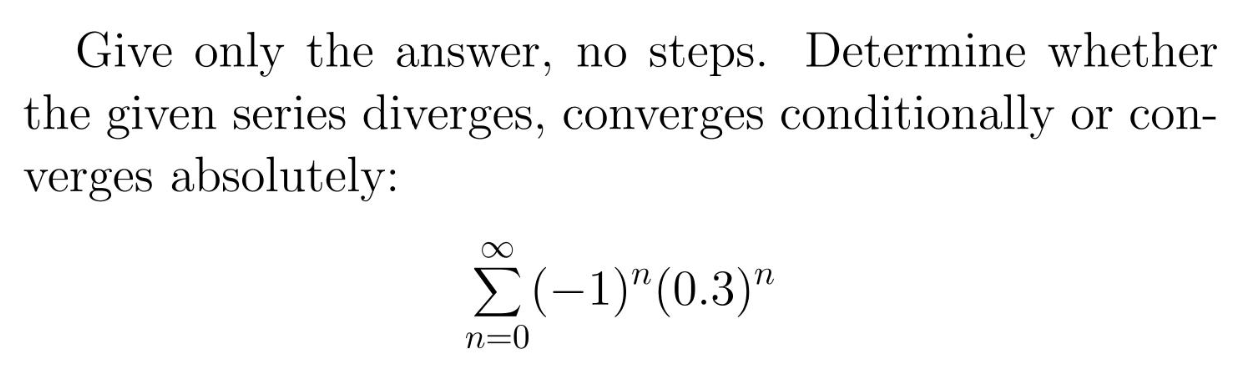}
    \caption{Sample 1 of Math Equation Solving (Hard) Dataset: Image.  }
    \label{fig:MESH1}
\end{figure*}

\begin{table*}[ht]
    \centering
    
    \resizebox{\textwidth}{!}
    {%
    \begin{tabular}{|lp{1\linewidth}|}

    \hline
      \textbf{Text:}&
\texttt Give only the answer, no steps. Determine whether the given series diverges, converges conditionally or converges absolutely:
      \\ &\$\$
        \textbackslash sum\_\{n=0\}\^ \ \{\textbackslash infty\}(-1)\^ \ n(0.3)\^ \ n\$\$
    \\ 
   \hline \end{tabular} 
    }

    \caption{Sample 1 of Math Equation Solving (Hard) Dataset: Text  
    }
    \label{table:MESH1}
    
\end{table*}

\clearpage

\begin{figure*}[ht]
    \centering
    \includegraphics[width=0.72\linewidth]{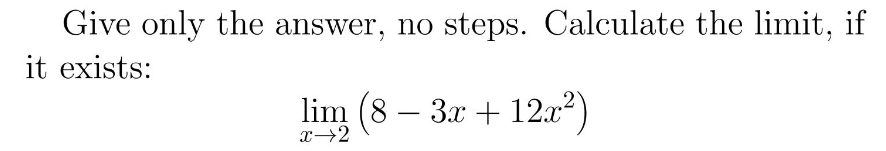}
    \caption{Sample 2 of Math Equation Solving (Hard) Dataset: Image.  }
    \label{fig:MESH2}
\end{figure*}

\begin{table*}[h]
    \centering
    
    \resizebox{0.95\textwidth}{!}
    {%
    \begin{tabular}{|lp{1\linewidth}|}

    \hline
      \textbf{Text:}&
\texttt  Give only the answer, no steps. Calculate the limit, if it exists:
\$\$ \textbackslash lim\_\{ x \textbackslash rightarrow 2 \} \textbackslash left (8-3 x+12 x\^ \ 2 \textbackslash right)\$\$
    \\ 
   \hline \end{tabular} 
    }

    \caption{Sample 2 of Math Equation Solving (Hard) Dataset: Text  
    }
    \label{table:MESH2}
    
\end{table*}

\newpage

\section{LogicQA Dataset}
\label{app:LogiQA}

\begin{figure*}[ht]
    \centering
    \includegraphics[width=0.7\linewidth]{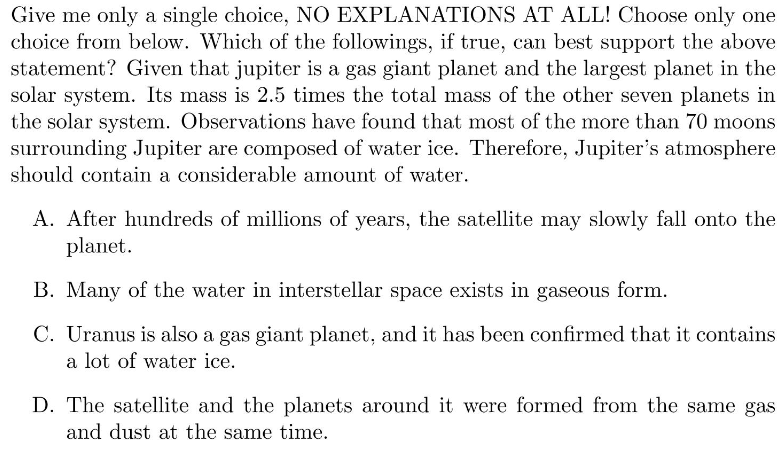}
    \caption{Sample 1 of LogicQA Dataset: Image.  }
    \label{fig:LogiQA1}
\end{figure*}

\begin{table*}[ht]
    \centering
    
    \resizebox{0.95\textwidth}{!}
    {%
    \begin{tabular}{|lp{1\linewidth}|}

    \hline
      \textbf{Text:}&Give me only a single choice, NO EXPLANATIONS AT ALL! Choose only one choice from below. Which of the followings, if true, can best support the above statement? Given that jupiter is a gas giant planet and the largest planet in the solar system. Its mass is 2.5 times the total mass of the other seven planets in the solar system. Observations have found that most of the more than 70 moons surrounding Jupiter are composed of water ice. Therefore, Jupiter's atmosphere should contain a considerable amount of water.
      
A. After hundreds of millions of years, the satellite may slowly fall onto the planet.

B. Many of the water in interstellar space exists in gaseous form.

C. Uranus is also a gas giant planet, and it has been confirmed that it contains a lot of water ice.

D. The satellite and the planets around it were formed from the same gas and dust at the same time.

    \\ 
   \hline \end{tabular} 
    }

    \caption{Sample 1 of LogicQA Dataset: Text  
    }
    \label{table:LogiQA1}
    
\end{table*}

\clearpage
\begin{figure*}[ht]
    \centering
    \includegraphics[width=0.7\linewidth]{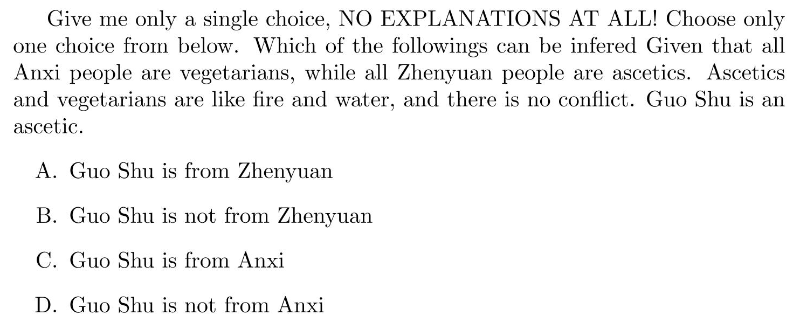}
    \caption{Sample 2 of LogicQA Dataset: Image.  }
    \label{fig:LogiQA2}
\end{figure*}

\begin{table*}[ht]
    \centering
    
    \resizebox{0.95\textwidth}{!}
    {%
    \begin{tabular}{|lp{1\linewidth}|}

    \hline
      \textbf{Text:}&Give me only a single choice, NO EXPLANATIONS AT ALL! Choose only one choice from below. Which of the followings can be infered Given that all Anxi people are vegetarians, while all Zhenyuan people are ascetics. Ascetics and vegetarians are like fire and water, and there is no conflict. Guo Shu is an ascetic.
      
A. Guo Shu is from Zhenyuan

B. Guo Shu is not from Zhenyuan

C. Guo Shu is from Anxi

D. Guo Shu is not from Anxi

    \\ 
    
   \hline \end{tabular} 
    }

    \caption{Sample 2 of LogicQA Dataset: Text  
    }
    \label{table:LogiQA2}
    
\end{table*}

\clearpage

\section{MMLU Dataset}
\label{app:MMLU}

\begin{figure*}[ht]
    \centering
    \includegraphics[width=0.7\linewidth]{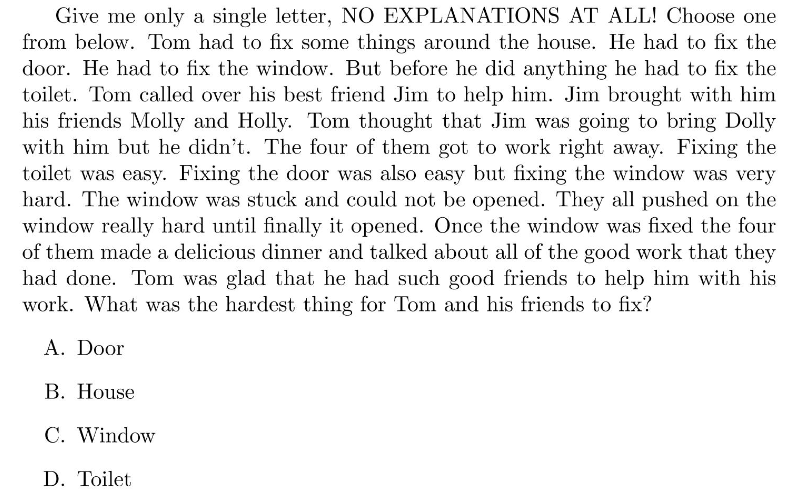}
    \caption{Sample 1 of MMLU Dataset: Image.  }
    \label{fig:MMLU1}
\end{figure*}

\begin{table*}[ht]
    \centering
    
    \resizebox{0.95\textwidth}{!}
    {%
    \begin{tabular}{|lp{1\linewidth}|}

    \hline
      \textbf{Text:}&Give me only a single letter, NO EXPLANATIONS AT ALL! Choose one from below. Tom had to fix some things around the house. He had to fix the door. He had to fix the window. But before he did anything he had to fix the toilet. Tom called over his best friend Jim to help him. Jim brought with him his friends Molly and Holly. Tom thought that Jim was going to bring Dolly with him but he didn't. The four of them got to work right away. Fixing the toilet was easy. Fixing the door was also easy but fixing the window was very hard. The window was stuck and could not be opened. They all pushed on the window really hard until finally it opened. Once the window was fixed the four of them made a delicious dinner and talked about all of the good work that they had done. Tom was glad that he had such good friends to help him with his work. What was the hardest thing for Tom and his friends to fix?
      
A. Door

B. House

C. Window

D. Toilet

    \\ 
   \hline \end{tabular} 
    }

    \caption{Sample 1 of MMLU Dataset: Text  
    }
    \label{table:MMLU1}
    
\end{table*}

\clearpage
\begin{figure*}[ht]
    \centering
    \includegraphics[width=0.7\linewidth]{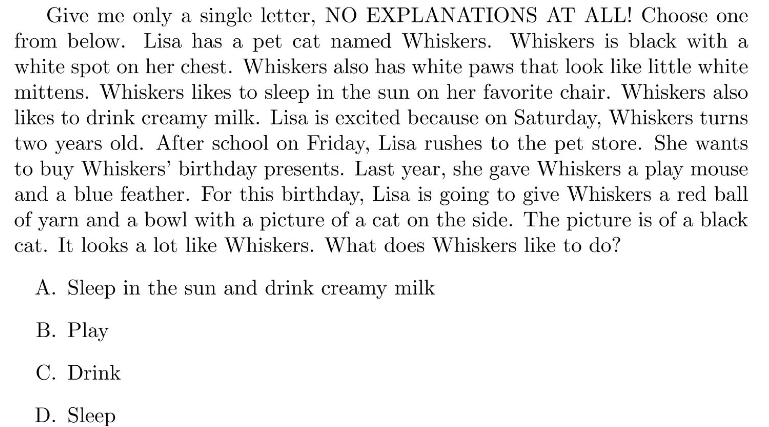}
    \caption{Sample 2 of MMLU Dataset: Image.  }
    \label{fig:MMLU2}
\end{figure*}

\begin{table*}[ht]
    \centering
    
    \resizebox{0.95\textwidth}{!}
    {%
    \begin{tabular}{|lp{1\linewidth}|}

    \hline
      \textbf{Text:}&Give me only a single letter, NO EXPLANATIONS AT ALL! Choose one from below. Lisa has a pet cat named Whiskers. Whiskers is black with a white spot on her chest. Whiskers also has white paws that look like little white mittens.  Whiskers likes to sleep in the sun on her favorite chair. Whiskers also likes to drink creamy milk.  Lisa is excited because on Saturday, Whiskers turns two years old.  After school on Friday, Lisa rushes to the pet store. She wants to buy Whiskers' birthday presents. Last year, she gave Whiskers a play mouse and a blue feather.  For this birthday, Lisa is going to give Whiskers a red ball of yarn and a bowl with a picture of a cat on the side. The picture is of a black cat. It looks a lot like Whiskers. What does Whiskers like to do?
      
A. Sleep in the sun and drink creamy milk

B. Play

C. Drink

D. Sleep

    \\ 
    
   \hline \end{tabular} 
    }

    \caption{Sample 2 of MMLU Dataset: Text  
    }
    \label{table:MMLU2}
    
\end{table*}

\clearpage

\section{Table Understanding Dataset}
\label{app:TU}

\begin{figure*}[ht]
    \centering
    \includegraphics[width=0.6\linewidth]{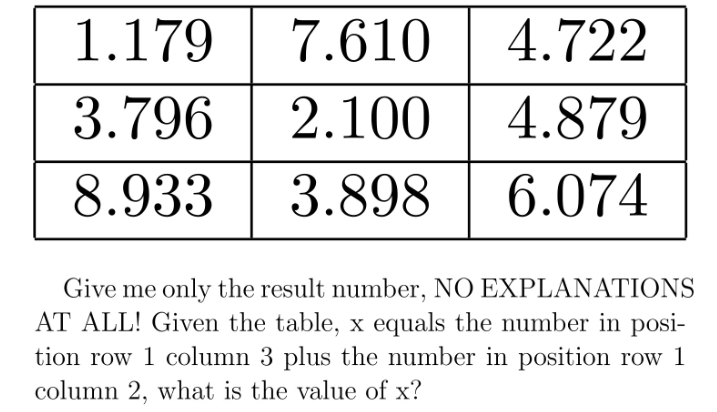}
    \caption{Sample 1 of Table Understanding Dataset: Image.  }
    \label{fig:TU1}
\end{figure*}

\begin{table*}[ht]
    \centering
    
    \resizebox{0.95\textwidth}{!}
    {%
    \begin{tabular}{|lp{1\linewidth}|}

    \hline
      \textbf{Text:}&Give me only the result number, NO EXPLANATIONS AT ALL! Given the table, x equals the number in position row 1 column 3 plus the number in position row 1 column 2, what is the value of x?
      
\textbackslash begin\{table\}[]

\textbackslash centering

\textbackslash resizebox\{\textbackslash textwidth\}\{!\}\{\%

\textbackslash begin\{tabular\}\{|l|l|l|\}

\textbackslash hline 

1.179 \& 7.610 \& 4.722  \textbackslash \textbackslash 

\textbackslash  hline

3.796 \& 2.100 \& 4.879 \textbackslash \textbackslash

\textbackslash hline 

8.933 \& 3.898 \& 6.074 \textbackslash 
\textbackslash

\textbackslash hline 

\textbackslash end\{tabular\}\}

\textbackslash end\{table\}

    \\ 
   \hline \end{tabular} 
    }

    \caption{Sample 1 of Table Understanding Dataset: Text  
    }
    \label{table:TU1}
    
\end{table*}

\clearpage
\begin{figure*}[ht]
    \centering
    \includegraphics[width=0.6\linewidth]{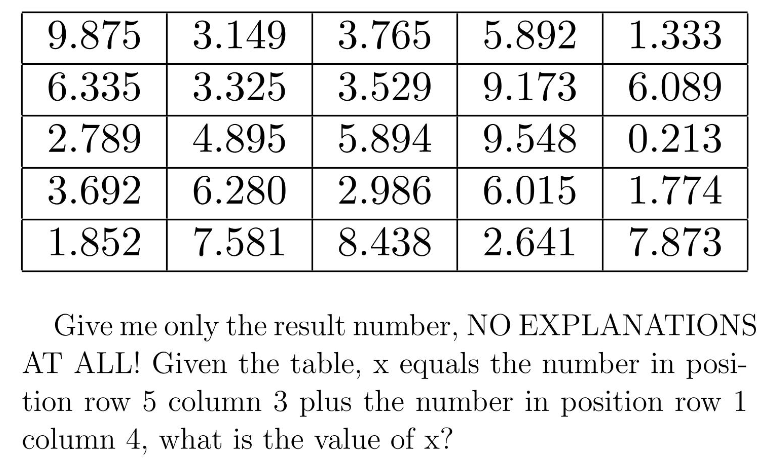}
    \caption{Sample 2 of Table Understanding Dataset: Image.  }
    \label{fig:TU2}
\end{figure*}

\begin{table*}[ht]
    \centering
    
    \resizebox{0.95\textwidth}{!}
    {%
    \begin{tabular}{|lp{1\linewidth}|}

    \hline
      \textbf{Text:}&Give me only the result number, NO EXPLANATIONS AT ALL! Given the table, x equals the number in position row 5 column 3 plus the number in position row 1 column 4, what is the value of x?
      
\textbackslash begin\{table\}[]

\textbackslash centering

\textbackslash resizebox\{\textbackslash textwidth\}\{!\}\{\%

\textbackslash begin\{tabular\}\{|l|l|l|l|l|\}

\textbackslash hline 

9.875 \& 3.149 \& 3.765 \& 5.892 \& 1.333  

\textbackslash \textbackslash 

\textbackslash hline 

6.335 \& 3.325 \& 3.529 \& 9.173 \& 6.089

\textbackslash \textbackslash

\textbackslash hline 

2.789 \& 4.895 \& 5.894 \& 9.548 \& 0.213  

\textbackslash \textbackslash

\textbackslash hline 

3.692 \& 6.280 \& 2.986 \& 6.015 \& 1.774  

\textbackslash \textbackslash

\textbackslash hline 

1.852 \& 7.581 \& 8.438 \& 2.641 \& 7.873  

\textbackslash \textbackslash

\textbackslash hline

\textbackslash end\{tabular\}\}

\textbackslash end\{table\}
    \\ 
   \hline \end{tabular} 
    }

    \caption{Sample 2 of Table Understanding Dataset: Text  
    }
    \label{table:TU2}
    
\end{table*}

\clearpage

\section{Math Reasoning Dataset}
\label{app:MR}
\begin{figure*}[ht]
    \centering
    \includegraphics[width=0.8\linewidth]{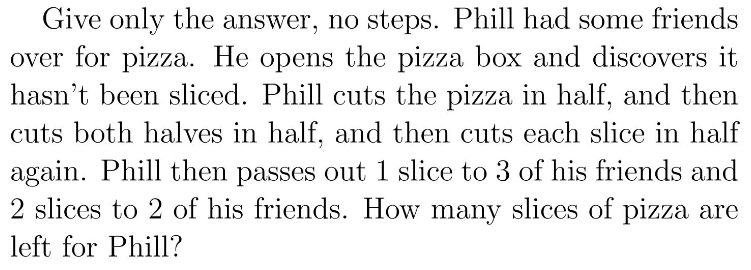}
    \caption{Sample 1 of Math Reasoning Dataset: Image.  }
    \label{fig:MR1}
\end{figure*}

\begin{table*}[ht]
    \centering
    
    \resizebox{0.85\textwidth}{!}
    {%
    \begin{tabular}{|lp{1\linewidth}|}

    \hline
      \textbf{Text:}&Give only the answer, no steps. Phill had some friends over for pizza.  He opens the pizza box and discovers it hasn't been sliced.  Phill cuts the pizza in half, and then cuts both halves in half, and then cuts each slice in half again.  Phill then passes out 1 slice to 3 of his friends and 2 slices to 2 of his friends.  How many slices of pizza are left for Phill?
    \\ 
   \hline \end{tabular} 
    }

    \caption{Sample 1 of Math Reasoning Dataset: Text  
    }
    \label{table:MR1}
    
\end{table*}

\clearpage
\begin{figure*}[ht]
    \centering
    \includegraphics[width=0.7\linewidth]{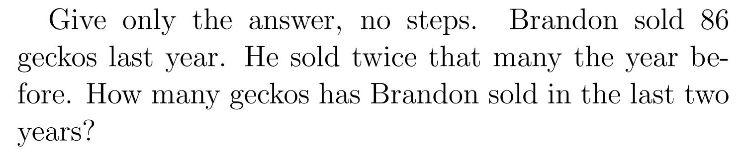}
    \caption{Sample 2 of Math Reasoning Dataset: Image.  }
    \label{fig:MR2}
\end{figure*}

\begin{table*}[ht]
    \centering
    
    \resizebox{0.95\textwidth}{!}
    {%
    \begin{tabular}{|lp{1\linewidth}|}

    \hline
      \textbf{Text:}&Give only the answer, no steps. Brandon sold 86 geckos last year. He sold twice that many the year before. How many geckos has Brandon sold in the last two years?
    \\ 
   \hline \end{tabular} 
    }

    \caption{Sample 2 of Math Reasoning Dataset: Text  
    }
    \label{table:MR2}
    
\end{table*}
\clearpage

\section{State Machine Dataset}
\label{app:state}

\begin{figure*}[ht]
    \centering
    \includegraphics[width=0.6\linewidth]{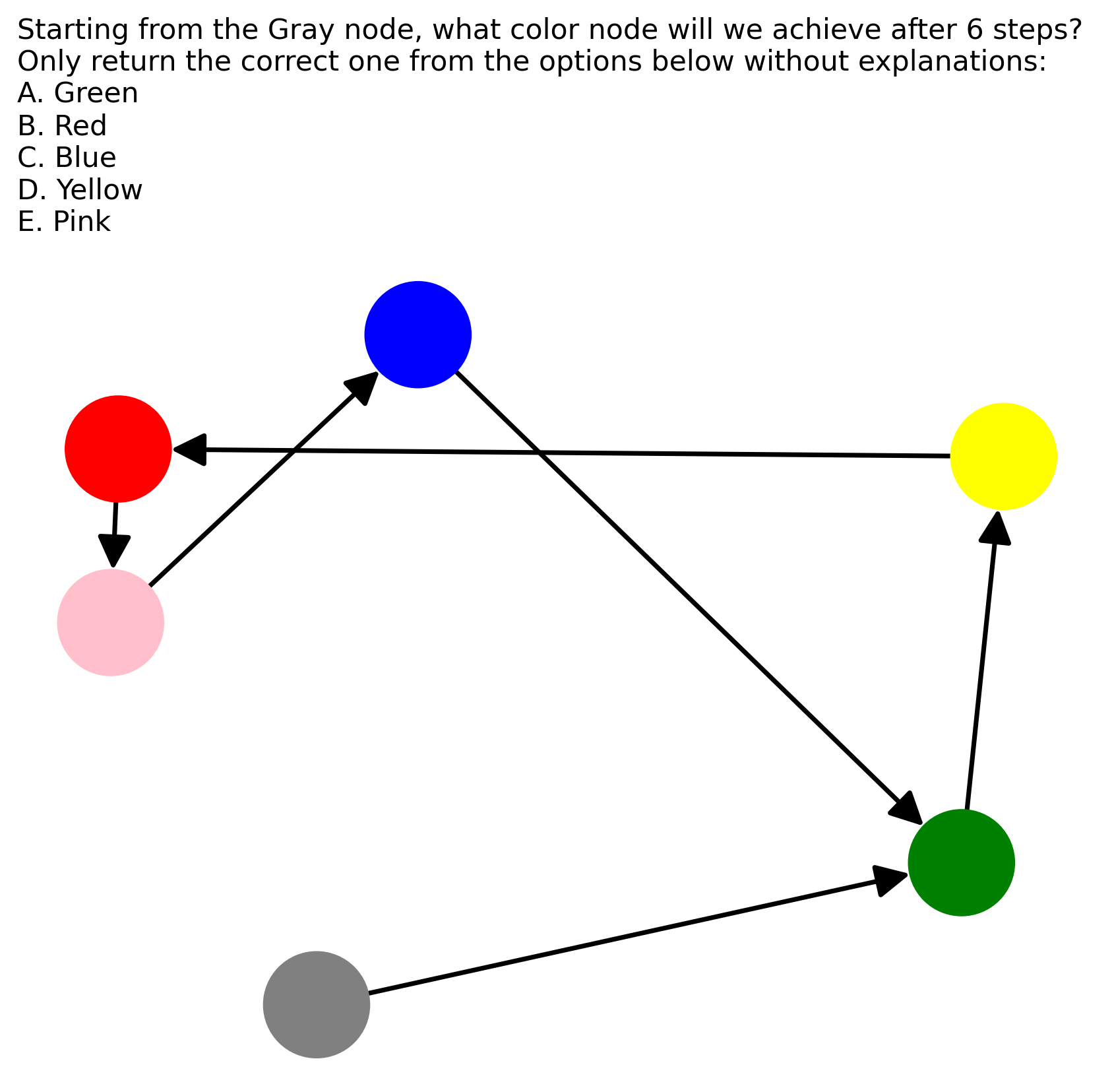}
    \caption{Sample 1 of State Machine Dataset: Image.  }
    \label{fig:SM1}
\end{figure*}

\begin{table*}[ht]
    \centering
    
    \resizebox{0.95\textwidth}{!}
    {%
    \begin{tabular}{|lp{1\linewidth}|}

    \hline
      \textbf{Text:}&Consider a graph with the following directed edges: 
Yellow leads to Red;
Green leads to Yellow;
Red leads to Pink;
Blue leads to Green;
Gray leads to Green;
Pink leads to Blue.
Starting from the Gray node, what color node will we achieve after 6 steps?
Only return the correct one from the options below without explanations:
A. Green
B. Red
C. Blue
D. Yellow
E. Pink
    \\ 
   \hline \end{tabular} 
    }

    \caption{Sample 1 of State Machine Dataset: Text  
    }
    \label{table:SM1}
    
\end{table*}

\clearpage
\begin{figure*}[ht]
    \centering
    \includegraphics[width=0.6\linewidth]{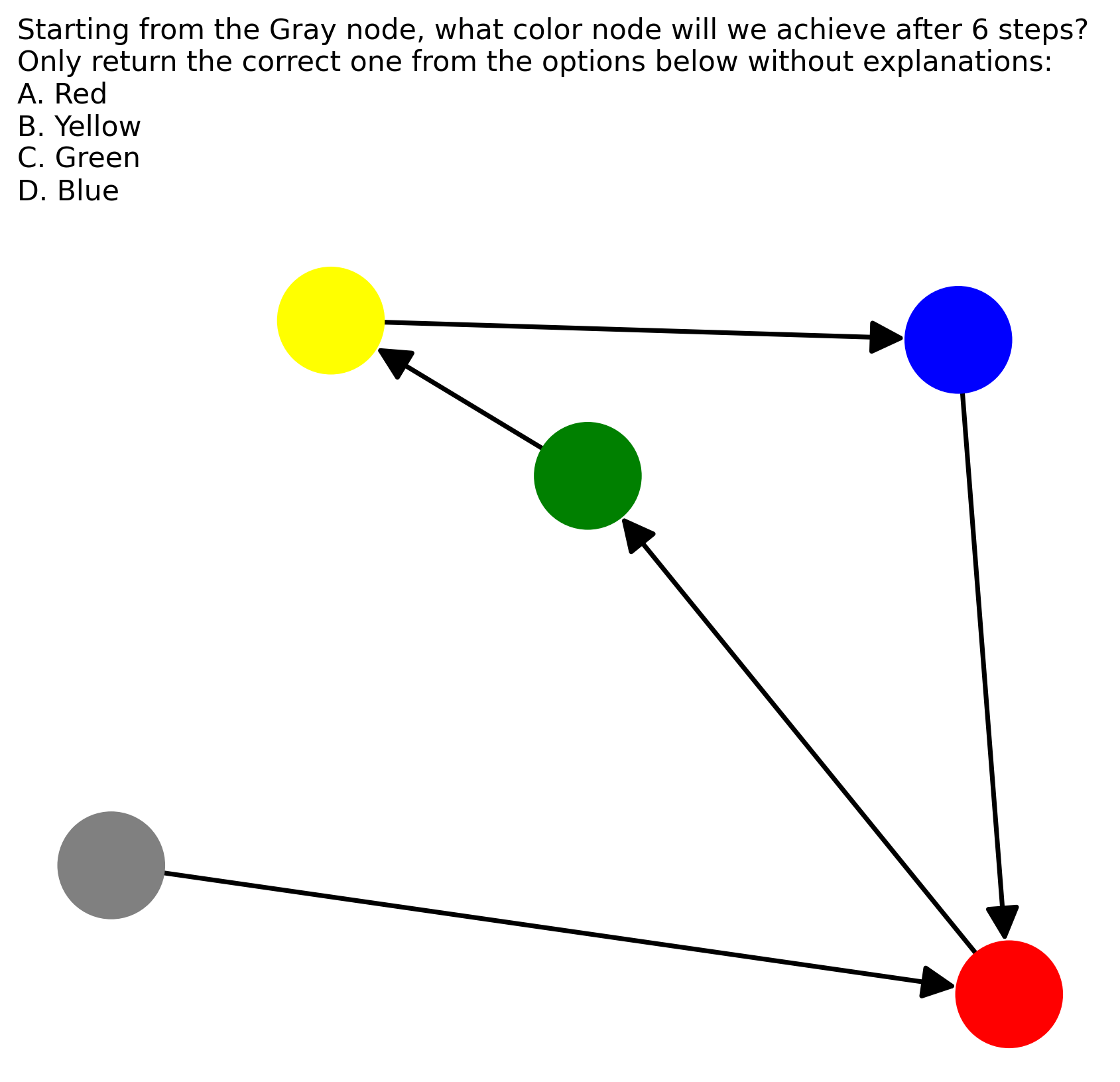}
    \caption{Sample 2 of State Machine Dataset: Image.  }
    \label{fig:SM2}
\end{figure*}

\begin{table*}[ht]
    \centering
    
    \resizebox{0.95\textwidth}{!}
    {%
    \begin{tabular}{|lp{1\linewidth}|}

    \hline
      \textbf{Text:}&Consider a graph with the following directed edges: 
Gray leads to Red;
Yellow leads to Blue;
Blue leads to Red;
Red leads to Green;
Green leads to Yellow.
Starting from the Gray node, what color node will we achieve after 6 steps?
Only return the correct one from the options below without explanations:
A. Red
B. Yellow
C. Green
D. Blue
    \\ 
   \hline \end{tabular} 
    }

    \caption{Sample 2 of State Machine Dataset: Text  
    }
    \label{table:SM2}
    
\end{table*}

\clearpage

\end{document}